\documentclass{article}
\usepackage{graphicx} % Required for inserting images
\usepackage[margin=2.5cm]{geometry} % 调整页边距以扩大横向布局
\usepackage[table]{xcolor} % Add xcolor package with table option
\usepackage{amsmath}
\usepackage{amssymb}
\usepackage{indentfirst}
\usepackage{microtype}
\usepackage{multirow}
\usepackage{booktabs}
\usepackage{longtable}
\usepackage{iclr_conference}

\usepackage{xltabular}
\usepackage{multirow}
\usepackage{booktabs}
\usepackage[table]{xcolor}
\iclrfinalcopy

\title{Agent-as-Tool: A Study on the Hierarchical Decision Making with Reinforcement Learning}
\author{%
  Yanfei Zhang \\
  Independent Researcher \\
  \texttt{zyf909533@gmail.com}
}

\begin{document}

\maketitle

\begin{abstract}
    \noindent
Large Language Models (LLMs) have emerged as one of the most significant technological advancements in artificial intelligence in recent years. Their ability to understand, generate, and reason with natural language has transformed how we interact with AI systems. With the development of LLM-based agents and reinforcement-learning-based reasoning models, the study of applying reinforcement learning in agent frameworks has become a new research focus. However, all previous studies face the challenge of deciding the tool calling process and the reasoning process simultaneously, and the chain of reasoning was solely relied on the unprocessed raw result with redundant information and symbols unrelated to the task from the tool, which impose a heavy burden on the model's capability to reason. Therefore, in our research, we proposed a hierarchical framework \textbf{Agent-as-tool} that detach the tool calling process and the reasoning process, which enables the model to focus on the verbally reasoning process while the tool calling process is handled by another agent. Our work had achieved comparable results with only a slight reinforcement fine-tuning on 180 samples, and had achieved exceptionally well performance in Bamboogle \citep{brown2020language} with \textbf{63.2\%} of exact match and \textbf{75.2\%} in cover exact match, exceeding Search-R1 by 4.8\% in exact match and 3.2\% in cover exact match. 
\end{abstract}

\section{Introduction}

Large Language Models (LLMs) have achieved remarkable progress in a wide range of natural language understanding and generation tasks\citep{liu2025advances, zhang2024aflow}. As the complexity of tasks increases, a common approach is to augment LLMs with access to external tools, such as web search engines, calculators, or code interpreters. This tool-augmented paradigm enables agents to interact with the environment and perform planning, reasoning, and execution steps beyond the model’s pretraining distribution.

Recent advancements have explored integrating reinforcement learning (RL) into these agent frameworks, aiming to improve decision-making over tool usage and multi-hop reasoning steps \citep{guo2025deepseek, jin2025searchr1trainingllmsreasoning}. However, a major limitation remains: existing RL-enhanced agents conflate the tool invocation process with the verbal reasoning process. This tight coupling leads to several challenges: (1) The agent must learn tool selection, input construction, and reasoning jointly, which increases training difficulty and noise; (2) Reasoning often proceeds over noisy, unstructured outputs returned directly from external tools, which degrades answer quality.

\begin{figure}[t]
    \centering
    \includegraphics[width=\textwidth]{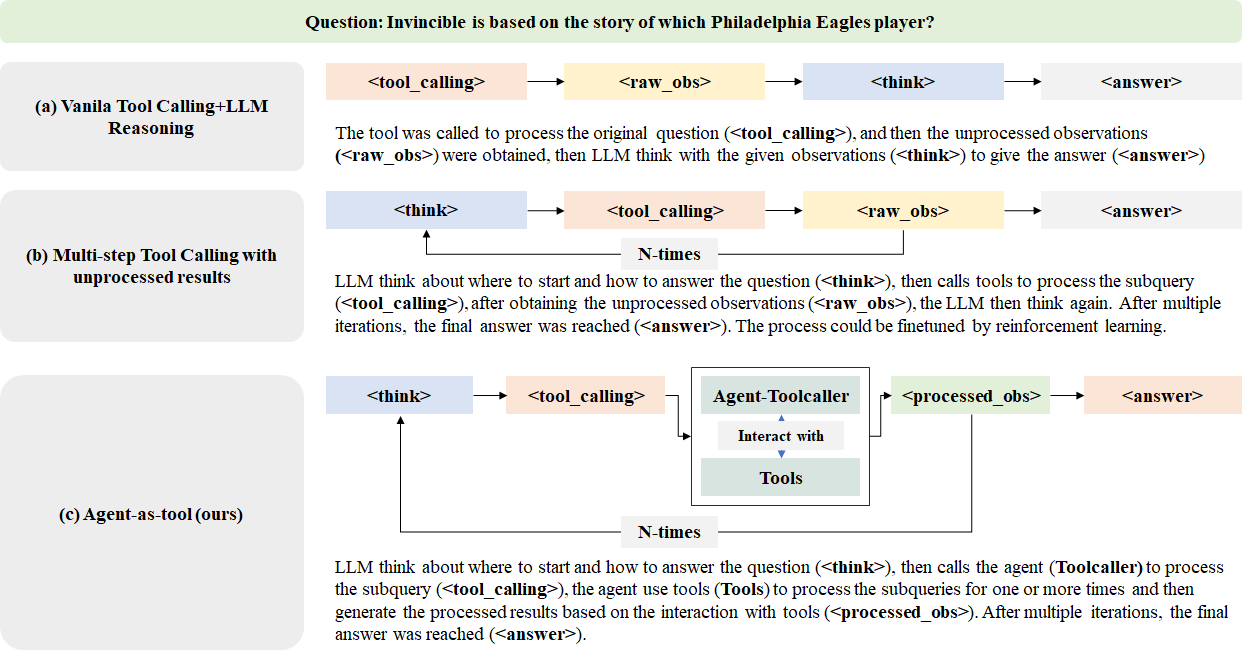}
    \caption{The trajectory of a single sample from a batch of questions processed in different research configurations. In our \textbf{Agent-as-tool} method, we employed the agent as a tool instead of calling the tool directly. The \textbf{Planner} is responsible for the tool calling process and the reasoning process, and the \textbf{Toolcaller} is responsible for the tool calling process to provide sufficient processed observations.}
    \label{fig:model_graph}
\end{figure}

To address these challenges, we propose \textbf{Agent-as-tool}, a hierarchical reasoning architecture in which reasoning and tool execution are explicitly decoupled as shown in Figure \ref{fig:model_graph}. The framework introduces a \textbf{Planner} and a \textbf{Toolcaller} as two separate agent components. The \textbf{Planner} focuses on natural language reasoning and high-level decision-making, while the \textbf{Toolcaller} is responsible for managing the tool interface (e.g., invoking web search) and returning structured observations.

The advantages of this design are twofold: (1) It simplifies the RL optimization process by assigning each sub-agent a focused objective; (2) It improves reasoning accuracy by allowing the \textbf{Planner} to operate on cleaner, more structured inputs. Furthermore, we apply a lightweight reinforcement fine-tuning procedure using GRPO on just 180 samples to demonstrate the efficiency of our framework.

This paper makes the following contributions:
\begin{itemize}
\item We propose \textbf{Agent-as-tool}, a hierarchical agent framework that separates reasoning and tool usage via a \textbf{Planner} and a \textbf{Toolcaller}.
\item We introduce a reinforcement learning protocol that enhances \textbf{Planner} behavior while masking \textbf{Toolcaller} outputs to preserve credit assignment integrity.
\item We empirically validate our framework on multiple multi-hop QA datasets and achieve state-of-the-art performance on Bamboogle.
\item We provide qualitative insights showing that hierarchical decoupling improves reasoning clarity and decomposition over existing baselines like Search-R1.
\end{itemize}

\section{Literature Review}\label{sec:literature_review}

\subsection{Agent Frameworks based on Pre-defined Reasoning Steps}

There are several agent researches that are designed to perform tasks with pre-trained LLMs and with pre-defined reasoning steps, including the CAMEL \citep{li2023camel}, OpenManus \citep{openmanus2025} and MetaGPT \citep{hong2023metagpt}. These works tend to extend the capabilities of the pre-trained LLM with additional rule-based reasoning steps to 'stimulate' the internal reasoning capabilities of the LLM to achieve better performances.

Specifically, considering the search and information retrieval for task completion scenario, there are also considerable works, including the Search-o1 \citep{li2025searcho1agenticsearchenhancedlarge}, OpenResearcher \citep{zheng2024openresearcher} (majorly focusing on the scientific research scenario).

\subsection{RL Reasoning Agents}

With the development of RL training frameworks and the Deepseek-R1 \citep{guo2025deepseek} setting, there are also considerable works to implement R1-style training paradigms on the LLM-based agents. The searching and information retrieval tasks were the first to be considered in this scenario, including the R1-searcher \citep{song2025r1} and DeepResearcher \citep{zheng2025deepresearcher}. 

There are also several works that integrate other external tools under the framework to complete different tasks, including the ToRL \citep{li2025torl} that integrate the python interpreter tool, ToolRL \citep{qian2025toolrl} that flexibly integrate different toolkits with different pre-defined datasets (e.g. API-Bank \citep{li2023apibankcomprehensivebenchmarktoolaugmented}), SWiRL \citep{goldie2025syntheticdatageneration} that control the tool selection process with different labels (\mbox{\texttt{<calculator>}} for calculator tool, and \mbox{\texttt{<search\_query>}} for web search tool). 

The generic process of calling an agent in these researches can be concluded as a sequence of thinking \mbox{\texttt{<think>}}, followed by a tool calling query enclosed with \mbox{\texttt{<tool\_query>}}, then the tool returns observations \mbox{\texttt{<obs>}}. With the reasoning on each step, the final answer could be reached whenever the agent think the ground truths are sufficient enough to give the final answer. It is a much simpler configuration with a ReAct-like tool calling process \citep{yao2023reactsynergizingreasoningacting}, then reinforcement learning are applied to explore whether the model could exhibit the capabilities beyond simple reasoning to reach the next hop, as shown in Figure \ref{fig:model_graph} as Multi-step Tool Calling with Unprocessed Results.

\section{Methodology}\label{sec:methodology}

We propose the \textbf{Agent-as-tool} framework as a hierarchical design for multi-hop reasoning tasks. It separates the planning and tool usage responsibilities between two agent components: a high-level \textbf{Planner} and a subordinate \textbf{Toolcaller}. The \textbf{Planner} manages reasoning and task decomposition, while the \textbf{Toolcaller} executes external actions such as web search. This section outlines the design of both components and the reinforcement learning procedure employed to optimize the \textbf{Planner}.

\subsection{Agent Architecture}

\subsubsection{Planner}
The \textbf{Planner} is a language model agent responsible for high-level reasoning and tool invocation decisions. It reasons about the current task state and emits tool usage instructions in natural language.

\textbf{Reasoning}: The \textbf{Planner} conducts internal reasoning enclosed in \texttt{<think>...</think>} tags, in line with DeepSeek-R1 conventions \citep{guo2025deepseek}. It uses previous observations and the original query to plan the next subtask.

\textbf{Tool Invocation}: Tool calls are expressed as sub-queries wrapped in \texttt{<tool\_calling>...</tool\_calling>} tags. These queries are interpreted by the \textbf{Toolcaller}, and the results are returned to the \textbf{Planner} as \texttt{<obs>...</obs>} blocks for further reasoning.

\subsubsection{Toolcaller}
The \textbf{Toolcaller} is a dedicated LLM-based agent designed to interface with external tools. In our implementation, it wraps a web search tool and processes queries issued by the \textbf{Planner}.

We implement the \textbf{Toolcaller} using a CAMEL-style chat agent \citep{li2023camel}, powered by GPT-4o-mini \citep{hurst2024gpt}. It could retrieve top-\emph{k} search results multiple times and returns structured summaries to the \textbf{Planner}. Although our current prototype uses only web search, the architecture supports extension to tools like calculators or code interpreters, also including MCP-based tool servers.

\subsection{Reinforcement Learning with GRPO}

\subsubsection{Training Objective}
We employ Generalized Reinforcement Policy Optimization (GRPO) \citep{shao2024deepseekmath} to fine-tune the \textbf{Planner}. The objective is:
\begin{equation}
\begin{split}
\mathcal{J}(\Theta) &= \mathbb{E}_{x \sim \mathcal{D}, \{y_i\}_{i=1}^G \sim \pi_{\text{old}}(\cdot | x)} \Bigg[ \frac{1}{G} \sum_{i=1}^G \Big[ \\
&\min\left(\frac{\pi_\Theta(y_i | x)}{\pi_{\text{old}}(y_i | x)}A_i, \text{clip}\left(\frac{\pi_\Theta(y_i | x)}{\pi_{\text{old}}(y_i | x)}, 1-\varepsilon, 1+\varepsilon\right)A_i\right) \\
&- \beta D_{\text{KL}}(\pi_\Theta || \pi_{\text{ref}}) \Big] \Bigg]
\end{split}
\end{equation}
where $x$ is sampled from dataset $\mathcal{D}$, $y_i$ is a rollout, $A_i$ is the advantage, $\varepsilon$ is the clipping threshold, and $\beta$ regulates KL penalty.

\subsubsection{Observation Masking}
To prevent reward leakage through Toolcaller-generated outputs, we mask the \texttt{<obs>} blocks during reward modeling and training. These segments are replaced with special token \texttt{<fim\_pad>}, which is trained to embed close to zero.

\subsubsection{Reward Function}
Our reward function balances correctness and formatting constraints:
\begin{equation}
\text{Reward} = \begin{cases}
    \text{F1 score} & \text{if answer is correctly formatted} \\
    -2 & \text{otherwise}
\end{cases}
\end{equation}
The model receives a high reward when generating a valid and correct response, and a penalty when output is malformed.

\section{Experiments}\label{sec:experiments}
\subsection{Experiment Settings}
\subsubsection{Model and Hyperparameters}
We use Qwen-2.5-7B-Instruct \citep{qwen2025qwen25technicalreport} as our base model. The training is conducted by an customized implementation of rollout and a customized implementation of GRPO on trl \citep{vonwerra2022trl}. At each training step, we sample a batch of training data from the training set and calculate the reward for each rollout. Then, we update the policy by maximizing the reward. 

The batch size is set to 3 for each training step and each sample contains 12 rollouts for each prompt. Each rollout contains at most 10 rounds of tool calling.

\subsubsection{Training Settings}

We conducted quite a small scale of training for the \textbf{Agent-as-tool}. We trained the \textbf{Agent-as-tool} for 60 steps with each step containing 3 training samples, and each training sample contains 12 rollouts, with total size of only 180 samples and 2160 rollouts. The training data entries were selected from the HotpotQA \citep{yang2018hotpotqa} and 2WikiMultiHopQA \citep{ho2020constructing} datasets with the same ratio as the R1-searcher \citep{song2025r1}.

During the training process, we observed that the loss of the \textbf{Agent-as-tool} is not stable for the first 30 steps, which is likely due to the small training data, but after 30 steps, the loss is stable and close to 0 and the performance of the \textbf{Agent-as-tool} also stabilized.

\begin{figure}[htb]
    \centering
    \includegraphics[width=\textwidth]{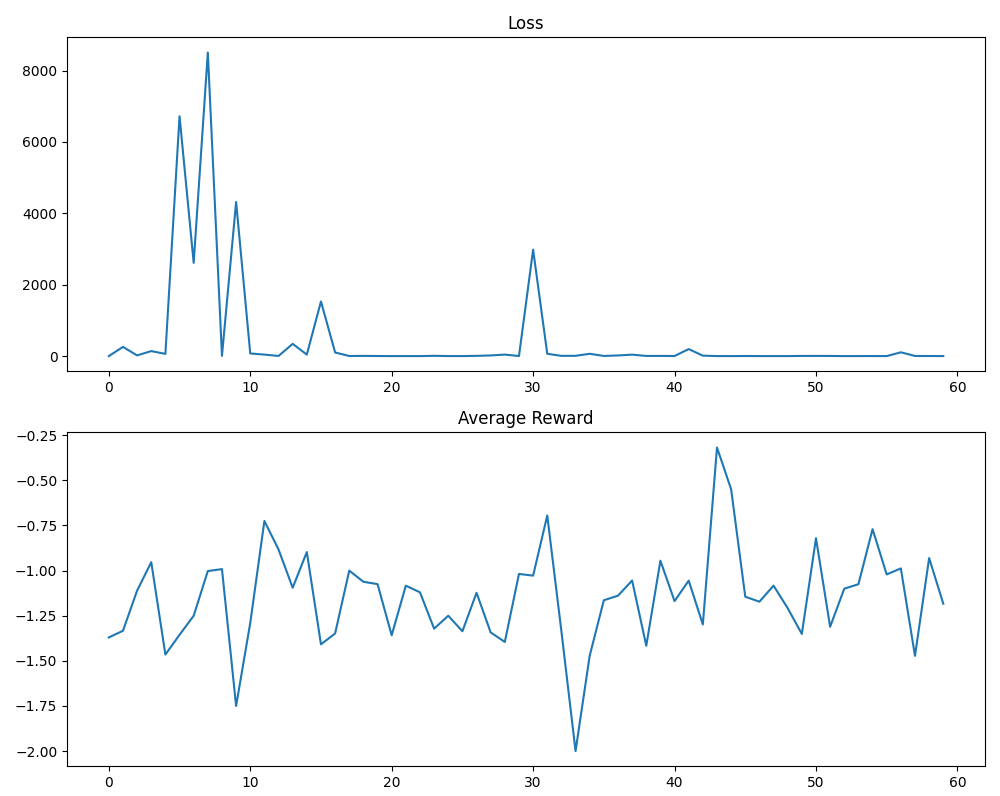}
    \caption{Training progress of the Agent-as-tool model showing loss convergence over training steps. The loss becomes stable after approximately 30 training steps.}
    \label{fig:training_graph}
\end{figure}
The training curve illustrated in Figure \ref{fig:training_graph} shows the convergence behavior of our model during the reinforcement learning process.

\subsubsection{Benchmark Settings}

In order to evaluate the performance of the \textbf{Agent-as-tool}, we conducted experiments on the open-domain question-answering task. We selected multiple multi-hop reasoning tasks to evaluate the performance of the \textbf{Agent-as-tool}, including the HotpotQA \citep{yang2018hotpotqa}, 2WikiMultiHopQA \citep{ho2020constructing}, MuSiQue \citep{trivedi2022musiquemultihopquestionssinglehop}, and bamboogle \citep{press2023measuringnarrowingcompositionalitygap}.

\subsubsection{Baseline Settings}

We have 1 information retrival tool: web search.

We then compare the performance of the \textbf{Agent-as-tool} with the following baselines:

\begin{itemize}
    \item \textbf{direct IO}: This baseline employs the direct output of the Qwen-2.5-7B-Instruct \citep{qwen2025qwen25technicalreport} as the answer without any external tool calling.
    \item \textbf{direct IO with web search}: This baseline employs the direct output of the Qwen-2.5-7B-Instruct \citep{qwen2025qwen25technicalreport}, but enables the web search to process the original question and return the top-k results as additional observations.
    \item \textbf{CAMEL Agent}: This baseline employs the CAMEL \citep{li2023camel} chat agent driven by the GPT-4o-mini \citep{hurst2024gpt}.
    \item \textbf{CAMEL Agent with web search}: This baseline employs the CAMEL \citep{li2023camel} chat agent driven by the GPT-4o-mini \citep{hurst2024gpt} and the same tool setting as the \textbf{Agent-as-tool} with web search tool only. This baseline is used as the reference for multi-hop reasoning tasks conducted with the rule-based agent framework.
    \item \textbf{Search-R1}: We directly compare the performance of the \textbf{Agent-as-tool} with the Search-R1 \citep{jin2025searchr1trainingllmsreasoning} in our configurations with web search tool for a fair comparison. As \textbf{Search-R1} cannot be directly integrated with the CAMEL \citep{li2023camel} chat agent, we would directly returns the search results as the answer instead of using another \textbf{Toolcaller}.
\end{itemize}

We conducted experiments with \textbf{Agent-as-tool} with pre-finetuned and post-finetuned models.

In align with the Deepseek-R1 setting \citep{guo2025deepseek}, we adopted the same prompt setting for all the baselines and the \textbf{Agent-as-tool} except Search-R1 \citep{jin2025searchr1trainingllmsreasoning} that is equipped with its orignal prompt setting, and we also modified the tool calling process to enable Search-R1 to accesss the unprocessed web search results.

\subsubsection{Evaluation Metrics}
In this paper, we focus on the performance of the \textbf{Agent-as-tool} in terms of the correctness of the answer, therefore, we employed the exact match metric (\textbf{EM}), the cover exact match metric (\textbf{CEM}) to evaluate the performance of the \textbf{Agent-as-tool}. 

\subsection{Quantitative Experiment Results}

The qualitative results are shown in \ref{tab:all_datasets_complete}. Based on the results, we can see that the \textbf{Agent-as-tool} outperforms most of the baselines except for the EM metric in the HotpotQA, 2WikiMultiHopQA, and MuSiQue datasets, where Search-R1 still has the best performance. However, in terms of the CEM metric, our model has a substantial improvement over all the baselines, except in HotpotQA where Search-R1 still has the best performance (64.2\% vs 57.4\%). And in the Bamboogle dataset \citep{press2023measuringnarrowingcompositionalitygap}, the \textbf{Agent-as-tool} with web search tool integrated to the \textbf{Toolcaller} (CAMEL \citep{li2023camel} agent) achieves the best performance with EM of 63.2\% and CEM of 75.2\%.

\begin{longtable}{clcc}
    \caption{Performance Comparison Across Different Datasets} \label{tab:all_datasets_complete} \\
    \toprule
    \textbf{Dataset} & \textbf{Model} & \textbf{EM (\%)} & \textbf{CEM (\%)}\\
    \midrule
    \endfirsthead
    
    \multicolumn{4}{c}%
    {\textbf{Table \thetable{} -- continued from previous page}} \\
    \toprule
    \textbf{Dataset} & \textbf{Model} & \textbf{EM (\%)} & \textbf{CEM (\%)}\\
    \midrule
    \endhead
    
    \midrule
    \multicolumn{4}{r}{\textit{Continued on next page}} \\
    \endfoot
    
    \bottomrule
    \endlastfoot
    
    \multirow{7}{*}{Bamboogle} 
    & Direct IO & 17.6 & 26.4 \\
    & Direct IO + Web Search & 29.6 & 42.4 \\
    & CAMEL & 36.8 & 47.2 \\
    & CAMEL + Web Search & 51.2 & 62.4 \\
    & Search-R1 + Web Search & 58.4 & 72.0 \\
    \cmidrule{2-4}
    \rowcolor{lightgray}
    & \textbf{Agent-as-tool-Base + Web Search} & 60.0 & 71.2 \\
    \rowcolor{lightgray}
    & \textbf{Agent-as-tool-Instruct + Web Search} & \textbf{63.2} & \textbf{75.2} \\
    \midrule
    
    \multirow{7}{*}{HotpotQA} 
    & Direct IO & 20.0 & 27.2 \\
    & Direct IO + Web Search & 32.6 & 52.8 \\
    & CAMEL & 23.2 & 44.2 \\
    & CAMEL + Web Search & 32.4 & 59.4 \\
    & Search-R1 + Web Search & \textbf{47.2} & \textbf{64.2} \\
    \cmidrule{2-4}
    \rowcolor{lightgray}
    & \textbf{Agent-as-tool-Base + Web Search} & 35.0 & 55.2 \\ 
    \rowcolor{lightgray}
    & \textbf{Agent-as-tool-Instruct + Web Search} & 37.2 & 57.4 \\
    \midrule
    
    \multirow{7}{*}{2WikiMultiHopQA} 
    & Direct IO & 22.6 & 25.4 \\
    & Direct IO + Web Search & 27.2 & 40.2 \\
    & CAMEL & 20.8 & 34.6 \\
    & CAMEL + Web Search & 35.0 & 69.4 \\
    & Search-R1 + Web Search & \textbf{52.4} & 68.0 \\
    \cmidrule{2-4}
    \rowcolor{lightgray}
    & \textbf{Agent-as-tool-Base + Web Search} & 42.8 & 68.0 \\ 
    \rowcolor{lightgray}
    & \textbf{Agent-as-tool-Instruct + Web Search} & 44.6 & \textbf{70.0} \\
    \midrule
    
    \multirow{7}{*}{MuSiQue} 
    & Direct IO & 4.8 & 9.0 \\
    & Direct IO + Web Search & 14.0 & 18.0 \\
    & CAMEL & 9.2 & 18.8 \\
    & CAMEL + Web Search & 16.0 & 29.4 \\
    & Search-R1 + Web Search & \textbf{20.8} & 28.6 \\
    \cmidrule{2-4}
    \rowcolor{lightgray}
    & \textbf{Agent-as-tool-Base + Web Search} & 15.6 & 28.8 \\ 
    \rowcolor{lightgray}
    & \textbf{Agent-as-tool-Instruct + Web Search} & 18.4 & \textbf{29.8} \\
    
\end{longtable}

We compared the performance of the \textbf{Agent-as-tool} before and after the reinforcement fine-tuning process. The table is shown in \ref{tab:finetune_improvements}. Based on the results, we can see that the Reinforcement fine-tuning based on GRPO \citep{shao2024deepseekmath} substantially improves the performance of the \textbf{Agent-as-tool} in all datasets with an average improvement of 2.5\% in EM and 2.3\% in CEM.

\begin{table}[htb]
    \centering
    \small
    \caption{Performance improvements after reinforcement fine-tuning}
    \resizebox{\textwidth}{!}{
        \begin{tabular}{l|cc|cc|cc}
        \hline
        \multirow{3}{*}{\textbf{Dataset}} & \multicolumn{2}{c|}{\textbf{Pre-finetuned}} & \multicolumn{2}{c|}{\textbf{Post-finetuned}} & \multicolumn{2}{c}{\textbf{Improvement}} \\
        \cline{2-7}
        & \textbf{EM} & \textbf{CEM} & \textbf{EM} & \textbf{CEM} & \textbf{EM} & \textbf{CEM} \\
        & (\%) & (\%) & (\%) & (\%) & (\%) & (\%) \\
        \hline
        Bamboogle & 60.0 & 71.2 & 63.2 & 75.2 & \textbf{+3.2} & \textbf{+4.0} \\
        HotpotQA & 35.0 & 55.2 & 37.2 & 57.4 & \textbf{+2.2} & \textbf{+2.2} \\
        2WikiMultiHopQA & 42.8 & 68.0 & 44.6 & 70.0 & \textbf{+1.8} & \textbf{+2.0} \\
        MuSiQue & 15.6 & 28.8 & 18.4 & 29.8 & \textbf{+2.8} & \textbf{+1.0} \\
        \hline
        \textbf{Average} & 38.4 & 55.8 & 40.9 & 58.1 & \textbf{+2.5} & \textbf{+2.3} \\
        \hline
        \end{tabular}
    }
    \label{tab:finetune_improvements}
\end{table}

Comparing with the CAMEL baseline with web search tool integrated (\textbf{CAMEL + Web Search}), the \textbf{Agent-as-tool} pre-finetuned and post-finetuned achieved a substantial improvement in EM and CEM, stating the necessity that the \textbf{Agent-as-tool} that enables the model to control when and what to be called in a tool calling is a more effective framework for multi-hop reasoning tasks.

Comparing with the Search-R1 baseline (\textbf{Search-R1 + Web Search}), which is the current best performing research of its kind, the \textbf{Agent-as-tool-Instruct} has substantial improvements over the Bamboogle dataset, which improves the EM by 4.8\% and CEM by 3.2\%, stating the effectiveness of the \textbf{Agent-as-tool} in multi-hop reasoning tasks. Besides, the \textbf{Agent-as-tool} conducted fine-tuning with 180 samples, which indicates the efficiency of the fine-tuning process.

\subsection{Qualitative Results Inspection and Analysis}

\subsubsection{Comparison of the Agent-as-tool and the Search-R1}

Comparing with Search-R1 baseline (\textbf{Search-R1 + Web Search}), the \textbf{Agent-as-tool-Instruct} had several advantages qualitatively:

\begin{itemize}
    \item The \textbf{Agent-as-tool-Instruct} could reason with less fuzzy and more structured observations, comparing with the \textbf{Search-R1 + Web Search} which would need to reason with the unprocessed web search results with fuzzy and unstructured symbols or other unrelated details.
    \item As the \textbf{Agent-as-tool-Instruct} adopt a hierarchical reasoning process which segragate the reasoning process and the tool calling process, the agent could have a better linearly text-based reasoning process comparing with the \textbf{Search-R1 + Web Search}.
\end{itemize}

The qualitative comparison as a example is shown in Figure \ref{fig:comparison_searchr1_and_ours} (in Appendix). The \textbf{Agent-as-tool-Instruct} could reason with less fuzzy and more structured observations, comparing with the \textbf{Search-R1 + Web Search} which would need to reason with the unprocessed web search results with fuzzy and unstructured symbols or other unrelated details.

\subsubsection{Comparison of the Results before and after the Reinforcement Fine-tuning}

Comparing with \textbf{Agent-as-tool-Base}, the \textbf{Agent-as-tool-Instruct} had several advantages qualitatively:

\begin{itemize}
    \item The \textbf{Agent-as-tool-Instruct} identify the correct decomposition of the question to identify the first hop and the second hop to be solved by the agent, comparing with the \textbf{Agent-as-tool-Base} which would not be able to decompose the multi-hop question correctly so it directly feed the agent with the whole question (only a sightly change from the original manner). If the agent is not capable of reasoning the multi-hop question correctly, the \textbf{Agent-as-tool-Base} would not be able to answer the question correctly.
    \item As the \textbf{Agent-as-tool-Instruct} was instructed to reason with the agent powered by the pretrained model, the fine-tuned model could give a more structured and reasonable question to be answered by the agent comparing with the \textbf{Agent-as-tool-Base}.
\end{itemize}

The qualitative comparison as a example is shown in Figure \ref{fig:comparison_before_and_after} (in Appendix). The \textbf{Agent-as-tool-Instruct} could correctly decompose the question to identify the first hop and the second hop, comparing with the \textbf{Agent-as-tool-Base} which would not be able to decompose the multi-hop question correctly so it directly feed the agent with the whole question.

\section{Conclusions and Future Work}\label{sec:conclusion}

\subsection{Conclusions}

In this paper, we majorly studied the multi-hop reasoning tasks with the \textbf{Agent-as-tool} framework. We found that the \textbf{Agent-as-tool} could achieve a substantial improvement in the performance of the multi-hop reasoning tasks, especially in the Bamboogle dataset \citep{press2023measuringnarrowingcompositionalitygap}. We also found that the \textbf{Agent-as-tool} could reason with less fuzzy and more structured observations, comparing with the \textbf{Search-R1 + Web Search} which would need to reason with the unprocessed web search results with fuzzy and unstructured symbols or other unrelated details.

\subsection{Limitations and Future Work}

This paper only assigns the search tool to the agent (or in another word, the search agent) so the scope is limited to the open-domain multi-hop search tasks. While because only 1 model was provided, the dynamic assignment of the tool to the agent is not considered. Therefore in our future work more tools would be considered to be assigned to the agent, while we would also explore the dynamic assignment of the tool to the agent, in another word, make the \textbf{Planner} as a \textbf{Tool Orchestrator}.

\pagebreak

% \printbibliography % Print the bibliography
\bibliographystyle{abbrvnat}
\bibliography{references}

\pagebreak

\appendix
\section{The qualitative Results inspection}

\subsection{The qualitative results inspection of the \textbf{Agent-as-tool-Instruct + Web Search} and the \textbf{Search-R1 + Web Search}}

\begin{figure}[htb]
    \centering
    \includegraphics[height=0.7\textheight]{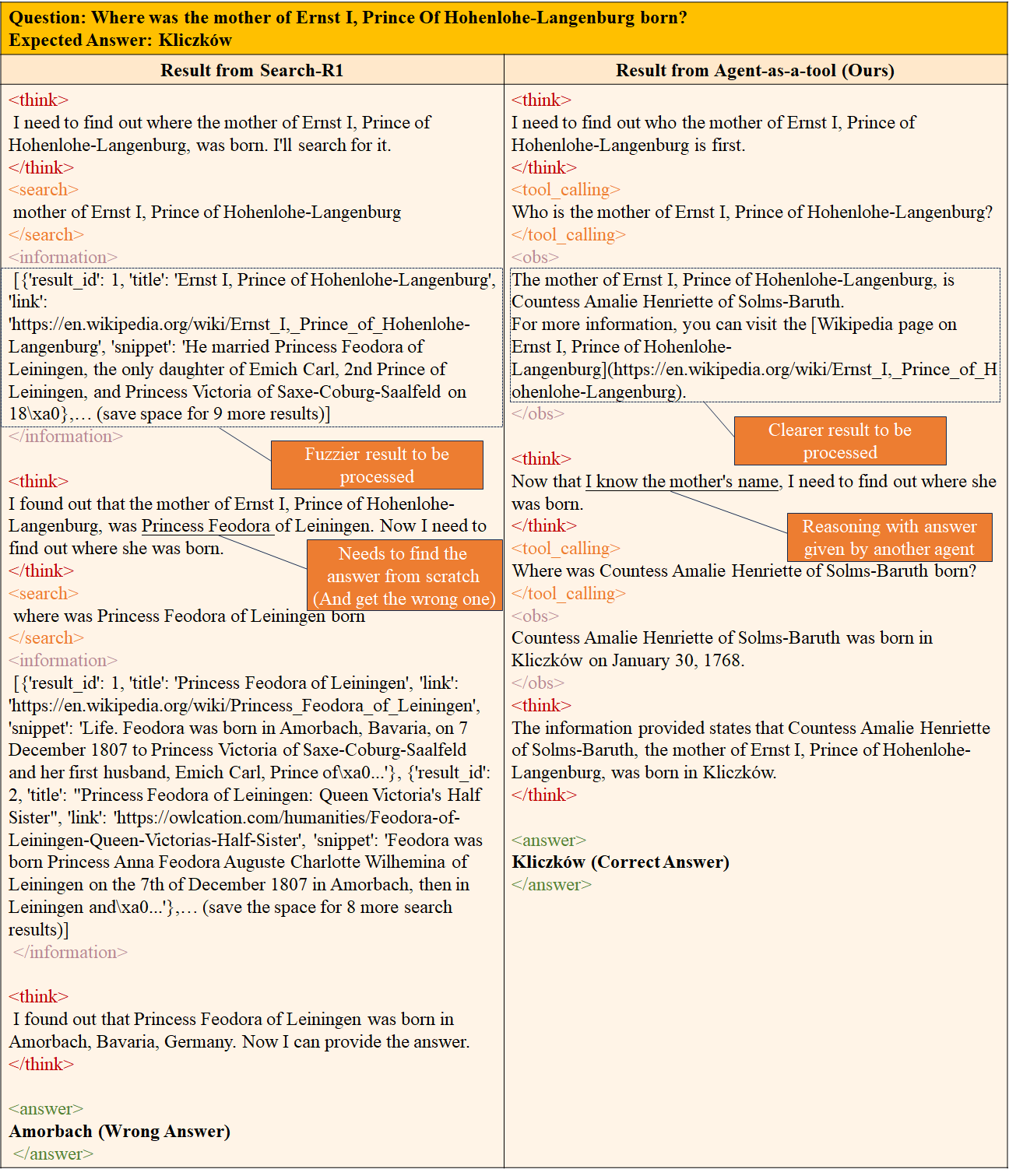}
    \caption{The \textbf{Agent-as-tool-Instruct} could reason with less fuzzy and more structured observations, comparing with the \textbf{Search-R1 + Web Search} which would need to reason with the unprocessed web search results with other unrelated details. Search-R1 was misled by the unprocessed web search results to reason with the wrong answer for the second hop (Princess Feodora of Leiningen), while the \textbf{Agent-as-tool-Instruct} has applied the agent to preprocess the web search results and return the correct answer (Countess Amalie Henriette of Solms-Baruth) for the second hop.}
    \label{fig:comparison_searchr1_and_ours}
\end{figure}

\pagebreak

\subsection{The qualitative results inspection of the \textbf{Agent-as-tool-Instruct + Web Search} and the \textbf{Agent-as-tool-Base + Web Search}}

\begin{figure}[htb]
    \centering
    \includegraphics[height=0.75\textheight]{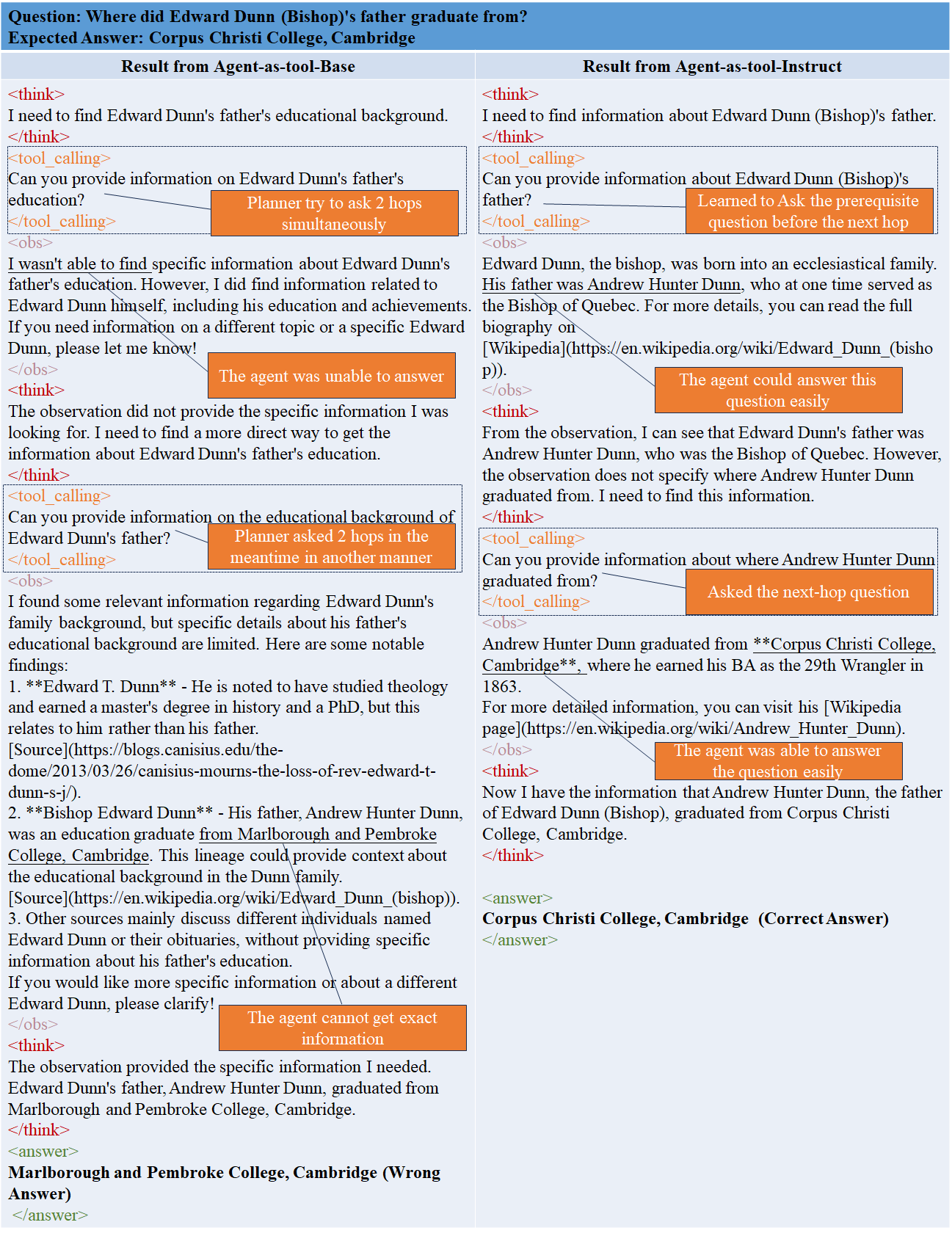}
    \caption{The \textbf{Agent-as-tool-Instruct + Web Search} could correctly decompose the question to identify the first hop and the second hop, comparing with the \textbf{Agent-as-tool-Base + Web Search} which barely decompose the question and try to ask about the whole question in another manner, i.e. the \textbf{Agent-as-tool-Base + Web Search} was not able to reason the whole multi-hop question correctly, while the \textbf{Agent-as-tool-Instruct + Web Search} could decompose the question to identify the first hop of the question to be answered by the agent then proceed to the second hop.}
    \label{fig:comparison_before_and_after}
\end{figure}

\end{document}